\documentclass[journal]{IEEEtran}
\usepackage{amsmath,amsfonts}
\usepackage{algorithmic}
\usepackage{algorithm}
\usepackage{array}
\usepackage[caption=false,font=normalsize,labelfont={rm,scriptsize},textfont={rm,scriptsize}]{subfig}
\usepackage{textcomp}
\usepackage{stfloats}
\usepackage{url}
\usepackage{verbatim}
\usepackage{graphicx}
\usepackage{cite}
\usepackage{bm}
\usepackage[utf8]{inputenc}
\usepackage{times}
\usepackage{enumitem}
\usepackage{balance}

\usepackage{booktabs}
\usepackage{xcolor}
\usepackage{pifont}

\usepackage{amssymb}
\usepackage{tabularx}
\usepackage{multirow}

\definecolor{mygreen}{rgb}{0,0.6,0}
\definecolor{myred}{rgb}{0.8,0,0}
\newcommand{\greencheck}{{\color{mygreen}\ding{51}}}%
\newcommand{\redcross}{{\color{myred}\ding{55}}}%

\newcommand{\pn}[1]{\textit{#1}}

\begin{document}
\bstctlcite{IEEEtran_bst_ctl}

\title{MoSKA: Mixture of Shared KV Attention for Efficient Long-Sequence LLM Inference}

\author{Myunghyun Rhee, Sookyung Choi, Euiseok Kim, Joonseop Sim, Youngpyo Joo, and Hoshik Kim
\thanks{Myunghyun Rhee, Sookyung Choi, Euiseok Kim, Joonseop Sim, Youngpyo Joo, and Hoshik Kim are with SK hynix Inc., Icheon-si, Gyeonggi-do 17336, Republic of Korea. (e-mail: \{myunghyun.rhee, sookyung.choi, euiseok.kim, joonseop.sim, youngpyo.joo, hoshik.kim\}@sk.com).}
}

\markboth{IEEE Journal of \LaTeX\ Class,~Vol.~12, No.~6, February~2024}%
{Shell \MakeLowercase{\textit{et al.}}: A Sample Article Using IEEEtran.cls for IEEE Journals}

\IEEEpubid{0000-0000~\copyright~2024 IEEE}

\maketitle
\begin{abstract}
The escalating context length in Large Language Models (LLMs) creates a severe performance bottleneck around the Key-Value (KV) cache, whose memory-bound nature leads to significant GPU under-utilization. This paper introduces \pn{Mixture of Shared KV Attention (MoSKA)}, an architecture that addresses this challenge by exploiting the heterogeneity of context data. It differentiates between per-request unique and massively reused shared sequences. The core of \pn{MoSKA} is a novel \pn{Shared KV Attention} mechanism that transforms the attention on shared data from a series of memory-bound GEMV operations into a single, compute-bound GEMM by batching concurrent requests. This is supported by an \pn{MoE-inspired sparse attention} strategy that prunes the search space and a tailored \pn{Disaggregated Infrastructure} that specializes hardware for unique and shared data.
This comprehensive approach demonstrates a throughput increase of up to 538.7× over baselines in workloads with high context sharing, offering a clear architectural path toward scalable LLM inference.
\end{abstract}

\begin{IEEEkeywords}
LLM Inference, Large Context, Shared KV Attention, Sparse Attention, Disaggregated Infrastructure.
\end{IEEEkeywords}

\section{Introduction}
\begin{figure}[!t]
\centering
\subfloat[Normalized KV cache size with optimizations.\label{fig:fig_1_a}]{%
\includegraphics[width=\columnwidth]{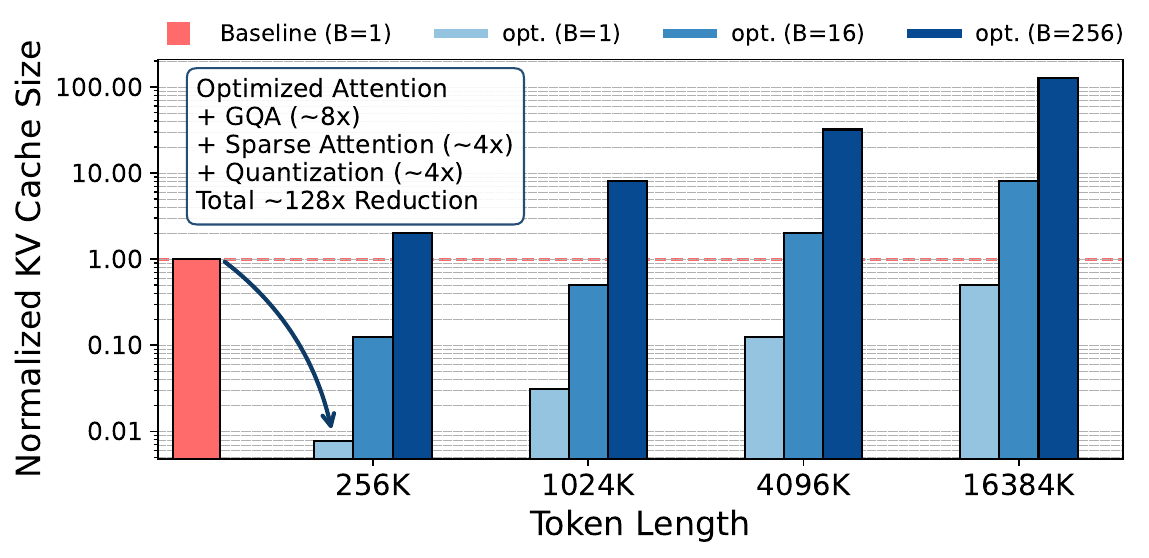}}
\hfil
\subfloat[Memory capacity and bandwidth requirement scaling with batch size.\label{fig:fig_1_b}]{%
\includegraphics[width=\columnwidth]{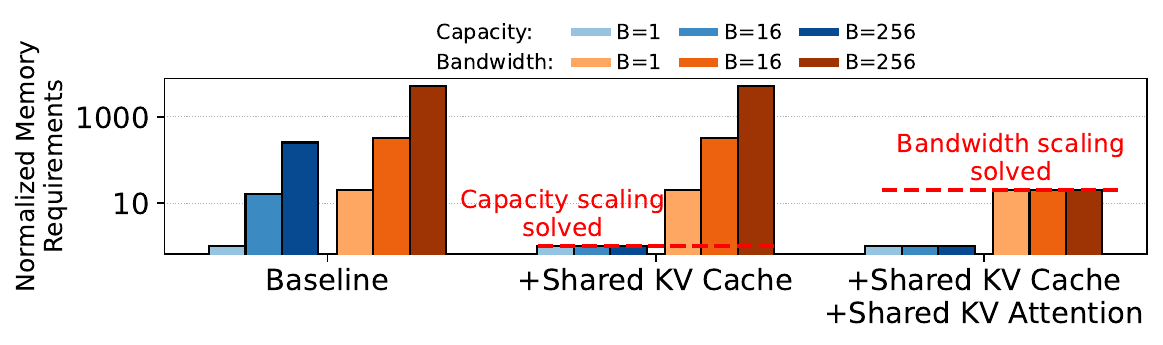}}
\caption{Hardware Requirement Challenges. (a) shows that even with significant optimizations (GQA, Sparse attention, Quantization) with widely-used optimization levels, KV cache size still scales with sequence length and batch size. (b) illustrates that while sharing the KV cache solves the memory capacity scaling, memory bandwidth requirements still scale with the batch size.
\pn{MoSKA}'s Shared KV Attention is designed to solve this remaining bandwidth scaling problem.
}
\label{fig:hw_requirements}
\vspace{-4mm}
\end{figure}

\IEEEPARstart{T}{he} evolution of transformer-based Large Language Models (LLMs) has entered an era defined by long context windows. While this progress enables sophisticated applications, it also exposes a fundamental performance bottleneck: the KV cache. As shown in Fig.~\ref{fig:hw_requirements}(a), even with a suite of modern optimizations such as Grouped-Query Attention (GQA), sparse attention, and quantization, the normalized KV cache size continues to scale with batch size and sequence length, posing a persistent challenge for system memory\cite{ChunkAttention2402}.

A more profound issue, however, lies in the scaling of memory bandwidth requirements. As depicted in Fig.~\ref{fig:hw_requirements}(b), simply sharing a KV cache among multiple requests can solve the memory \textit{capacity} scaling problem. However, the memory \textit{bandwidth} requirement still grows with the batch size. This is because each request's query must individually access the shared data, resulting in numerous memory-bound operations that underutilize the GPU's computational power. This points to a more fundamental challenge: solving the memory capacity issue alone is insufficient. A fundamental change in the attention computation itself is required.

\textcolor{black}{
To address this fundamental challenge, we leverage the heterogeneous nature of context data—distinguishing between \textit{unique}, per-request data and \textit{shared}, massively reused data. We introduce \pn{Mixture of Shared KV Attention (MoSKA)}, an architecture built around a novel \pn{Shared KV Attention} mechanism. This core component, combined with MoE-inspired sparsity and a tailored Disaggregated Infrastructure, resolves the critical bandwidth scaling bottleneck (Fig.~\ref{fig:hw_requirements}(b)) and enables truly scalable long-sequence LLM inference.
}

\IEEEpubidadjcol

\vspace{-2mm}
\section{Related Works and Motivation}
\renewcommand{\thesubsection}{\Alph{subsection}} 
\pn{MoSKA} builds upon the insights and limitations of several key research directions, weaving them into a cohesive architectural solution.

\begin{figure*}[t]
\centering
\subfloat[\pn{Shared KV Attention} vs. \pn{Unique KV Attention}.\label{fig_2_a}]{%
\includegraphics[width=0.95\columnwidth]{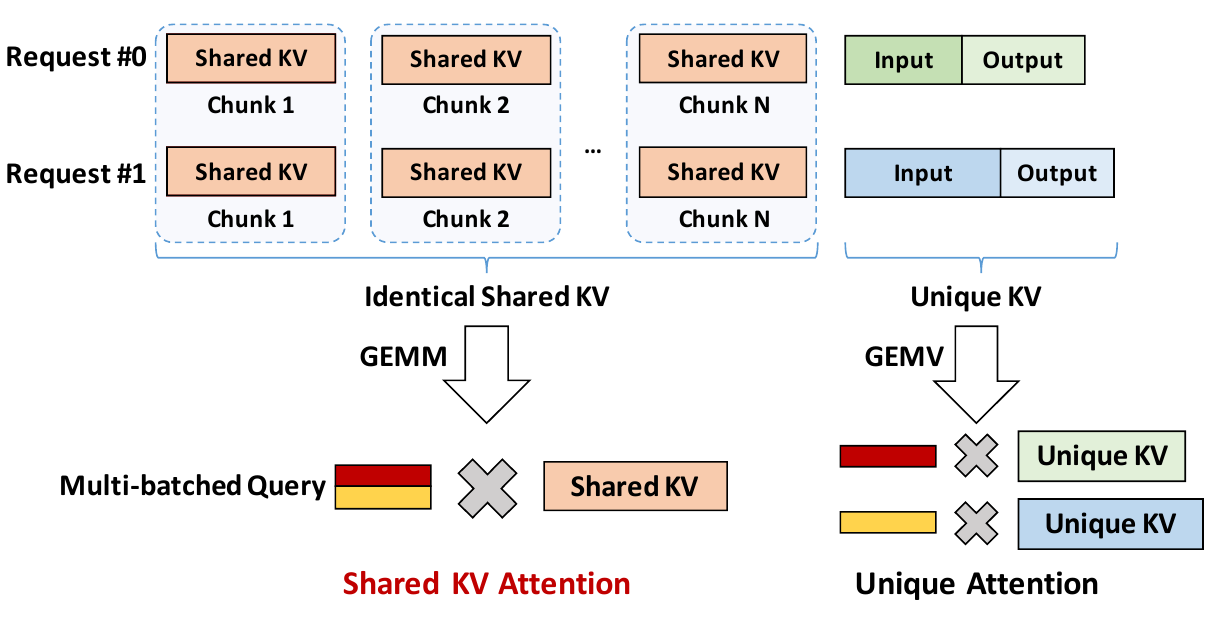}}
\hfil
\subfloat[\pn{MoSKA} architecture overview.\label{fig_2_b}]{%
\includegraphics[width=1.05\columnwidth]{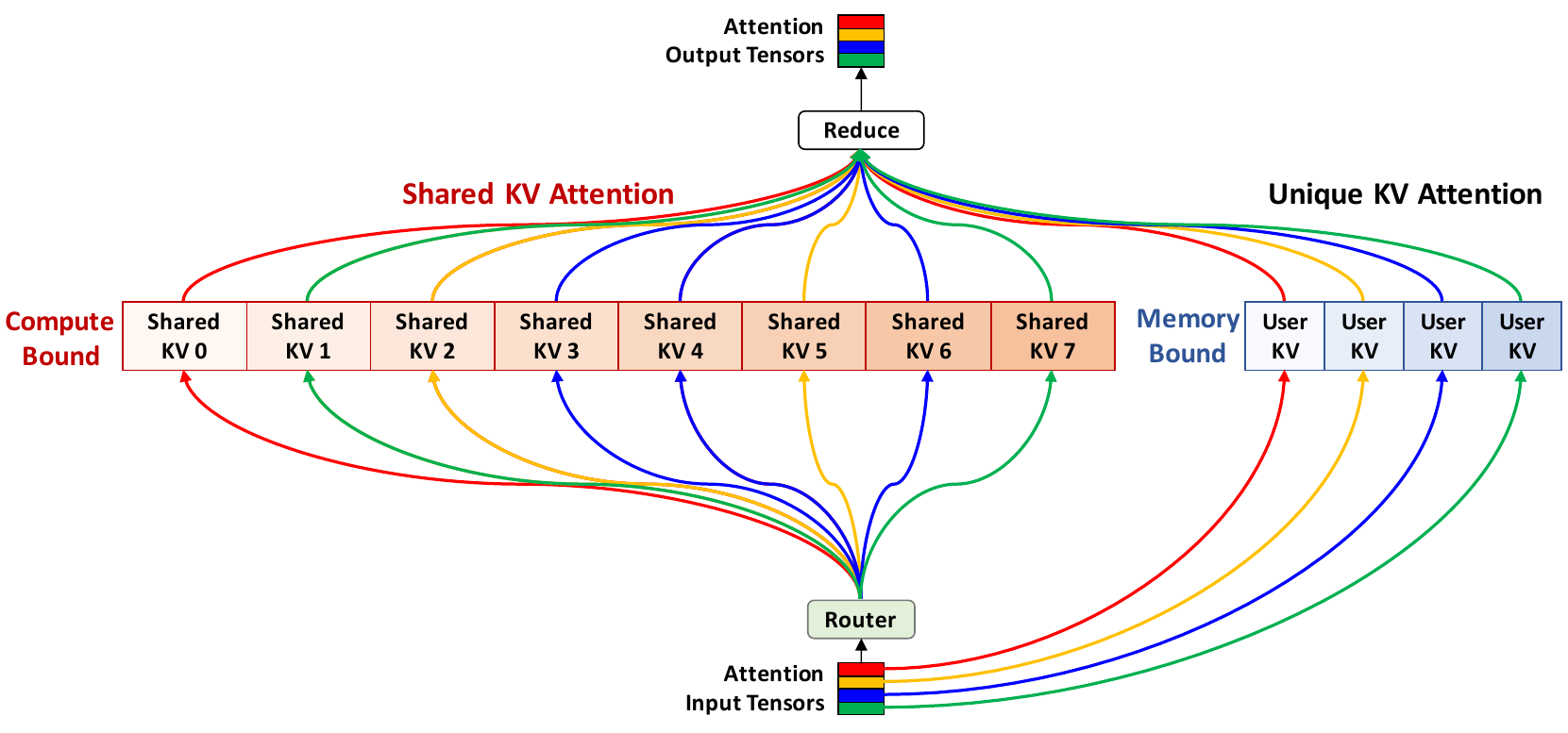}}
\caption{The \pn{MoSKA} Architecture.  detailing its core mechanism and high-level structure. (a) illustrates the fundamental principle of Shared KV Attention: concurrent queries to identical shared data are batched into a single, compute-bound GEMM operation, contrasting with the memory-bound GEMV operations used for unique KV data. Building on this, (b) depicts the complete \pn{MoSKA} system, where an MoE-inspired router first selects a sparse subset of relevant shared KV chunks (`Experts'), ensuring both computational efficiency and scalability over massive shared contexts.}
\label{fig:moska_arch}
\vspace{-4mm}
\end{figure*}

\vspace{-2mm}
\subsection{KV Cache Reuse}
The fundamental concept of improving inference efficiency via caching is well-established. The most direct application is \textit{prefix reuse}, where the KV cache from a previous turn in a multi-turn conversation is reused for the subsequent turn, avoiding redundant computation on the shared conversational history \cite{PromptCache2311}. This idea extends to scenarios where multiple, independent requests share a common prefix, such as a system prompt. Beyond simple prefix reuse, the paradigm of augmenting LLMs with vast, external knowledge bases, exemplified by Cache-Augmented Generation (CAG) \cite{DontDoRAG2412}, highlights a significant opportunity. These approaches motivate a more ambitious form of caching: pre-computing and maintaining the KV states of entire domain-specific documents (e.g., laws, medical cases) as persistent, shareable assets. This creates the potential for highly efficient, direct attention over massive shared contexts, but necessitates a sophisticated architecture to manage and serve these KV stores effectively.

\vspace{-1mm}

\subsection{Shared Attention, and System Disaggregation}
While powerful in concept, existing KV cache sharing implementations like \pn{SGLang} \cite{SGLang2312} and \pn{ChunkAttention} \cite{ChunkAttention2402} often rely on strict sequential prefix matching. This rigidity complicates batching in dynamic environments, failing to solve the bandwidth scaling problem (Fig.~\ref{fig:hw_requirements}(b)). Systems like \pn{FlashInfer} \cite{FlashInfer2501} also employ batching for shared prefixes, but their application is confined to contiguous prefix data.
\pn{MoSKA} distinguishes itself by enabling flexible batching of any identical shared data chunk, regardless of its position. This is a far more general approach, allowing \pn{MoSKA} to batch queries to shared data beyond simple prefixes, such as common legal clauses in documents or boilerplate code in programs. This flexibility significantly broadens the application scope of compute-bound GEMM conversion.

This flexibility in batching is complemented by a strategy to manage the complexity of attending to large contexts, MoE-inspired sparse attention has emerged as a critical technology.
Architectures like \pn{LongHeads} \cite{LongHeads2402} successfully use lightweight, training-free routers, while others like \pn{MoBA} \cite{MoBA2502} employ efficient non-parametric routers, to direct queries to only the most relevant data chunks.

Furthermore, while prior works like \pn{Lamina} \cite{Lamina2405} and \pn{MegaScale-Infer} \cite{MegaScaleInfer2504} explored attention/FFN offloading, they typically perform a general, layer-wise separation.
In contrast, \pn{MoSKA} introduces a more granular, attention-centric disaggregation. \pn{MoSKA} partitions the attention computation itself based on the data it processes. It specializes hardware for the distinct computational profiles of memory-bound operations on unique data versus compute-bound operations on shared data. This targeted separation by attention type is a more fundamental approach to resolving resource contention.

\vspace{-1mm}
\subsection{Positional Independent Transformer}
A fundamental constraint of most existing attention optimizations is the transformer's inherent dependence on positional embeddings. This dependency largely restricts KV cache reuse to contiguous and correctly ordered sequences. Recent, groundbreaking research into position-independent caching, such as EPIC \cite{EPIC2410}, charts a path to overcome this limitation. This strongly motivates the long-term \pn{Universal MoSKA} vision, where shared KV chunks are no longer tied to a specific context but exist as a modular, composable library of knowledge, which can be combined on-demand to service complex queries.

\vspace{-2mm}
\section{Proposed Architecture}
\renewcommand{\thesubsection}{\Alph{subsection}}
The \pn{MoSKA} architecture provides a systematic approach to processing large-scale shared KV data by managing it as a core, persistent resource within the system. Its constituent parts are carefully orchestrated to achieve significant, concurrent optimizations in memory, computation, and throughput.

\begin{table*}[!t]
\centering
\caption{Comparison of Key Features in Related Works and \pn{MoSKA}}
\label{tab:tech_comparison}
\begin{tabular}{l c c c c c}
\toprule
\textbf{} & \textbf{KV Reuse}& \textbf{Shared KV Attention} & \textbf{KV Routing} & \textbf{Disaggregated Infra.} & \textbf{Composable Context} \\
\midrule
FlashAttention \cite{Dao2022FlashAttentionFA}           & \redcross & \redcross & \redcross & \redcross & \redcross \\
SGLang  \cite{SGLang2312}                               & \greencheck & \redcross & \redcross & \redcross & \redcross \\
ChunkAttention, BatchLLM \cite{ChunkAttention2402, BatchLLM2412}                & \greencheck& \greencheck & \redcross & \redcross & \redcross  \\
LongHeads / MoBA \cite{LongHeads2402, MoBA2502}         & \redcross & \redcross & \greencheck & \redcross & \redcross \\
\midrule
\textbf{\pn{MoSKA}}                                     & \greencheck & \greencheck & \greencheck & \greencheck & \redcross\\
\textbf{\pn{Universal MoSKA}}                           & \greencheck & \greencheck & \greencheck & \greencheck & \greencheck\\
\bottomrule
\end{tabular}
\vspace{-3mm}
\end{table*}

\vspace{-2mm}
\subsection{The Shared KV Attention Mechanism}
The foundational principle of \pn{MoSKA} is to differentiate the processing of unique and shared data. As illustrated in Fig.~\ref{fig:moska_arch}(a), standard attention on per-request \pn{Unique KV} data remains a memory-bound GEMV operation. In contrast, for \pn{Identical Shared KV} data accessed by multiple requests, \pn{MoSKA} employs \pn{Shared KV Attention}. This mechanism aggregates $N$ concurrent queries into a multi-batched query matrix and processes them in a single, large-scale GEMM operation. This aggregation fundamentally shifts the operational bottleneck from memory bandwidth to computation by drastically increasing arithmetic intensity. The result is a substantial boost in GPU utilization and overall system throughput, directly addressing the bandwidth scaling problem identified in Fig.~\ref{fig:hw_requirements}(b). This operates on large, pre-computed \pn{Domain-Specific Shared KV Caches} that are managed with efficient chunking strategies.

\vspace{-2mm}
\subsection{The MoSKA Architecture}
While \pn{Shared KV Attention} transforms the core operation into a compute-bound GEMM, applying it naively across a massive, multi-million token shared KV cache would still be computationally prohibitive. To manage this complexity, \pn{MoSKA} incorporates an MoE-inspired routing layer, as shown in the architectural overview in Fig.~\ref{fig:moska_arch}(b). The entire shared KV space is first partitioned into smaller, manageable chunks, akin to 'experts' in an MoE model. For each incoming query, a lightweight, training-free routing mechanism dynamically calculates relevance scores and selects a top-k subset of the most pertinent shared KV chunks. 
To achieve this without a dedicated, trainable network, \pn{MoSKA} adopts a standard technique from prior lightweight MoE works\cite{LongHeads2402,MoBA2502}. It calculates semantic similarity via the inner product between the query and precomputed chunk embeddings.

This two-stage process is critical: the computationally intensive \pn{Shared KV Attention} is then executed only on this small, pre-selected subset of data. This approach achieves a powerful dual optimization. First, the routing acts as a sparse attention mechanism, drastically pruning the overall computational space and avoiding unnecessary work on irrelevant data. Second, the subsequent batched attention operation on the selected chunks still benefits from a high degree of batching, ensuring it remains a compute-bound task that maximizes hardware efficiency. This combination allows \pn{MoSKA} to efficiently query vast knowledge stores without incurring overwhelming computational costs.

\begin{figure}[t]
\centering
\includegraphics[width=\columnwidth]{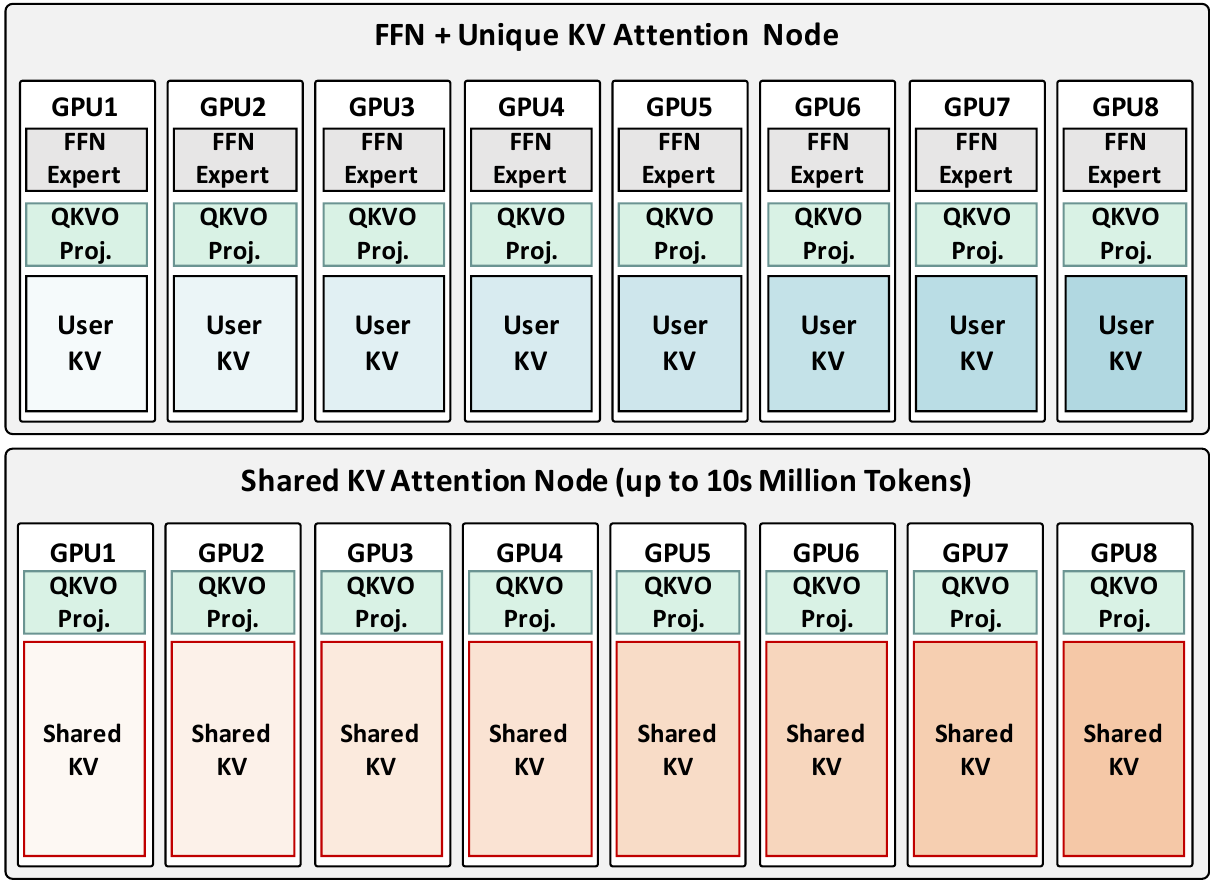}
\caption{Proposed disaggregated LLM serving infrastructure for \pn{MoSKA}, separating \pn{FFN}/\pn{Unique KV Attention} Nodes from specialized \pn{Shared KV Attention} Nodes.}
\label{fig:disaggregated_infra}
\vspace{-4mm}
\end{figure}

\vspace{-2mm}
\subsection{Disaggregated Infrastructure}
To fully realize the benefits of \pn{MoSKA}, we propose a Disaggregated Serving Infrastructure where hardware is specialized for the distinct computational profiles of unique and shared data (Fig.~\ref{fig:disaggregated_infra}). This avoids the resource contention and suboptimal performance of a monolithic system.

The infrastructure is split into two specialized node types. \pn{Unique KV} Nodes are optimized for latency-sensitive unique sequences, which are memory-bound (GEMV-based). These nodes utilize high-bandwidth memory and improve GPU utilization by co-locating the compute-intensive FFN layers to hide the memory latency of attention. In contrast, \pn{Shared KV} Nodes are designed for the throughput-oriented, compute-bound \pn{Shared KV Attention} task. They are equipped with powerful compute units (e.g., more GEMM cores) to efficiently process large batches. This specialization allows resources to be scaled independently, enabling a system to scale up its shared knowledge processing capacity without over-provisioning expensive, latency-optimized unique processing nodes.

\vspace{-2mm}
\subsection{Universal MoSKA for Composable Context}
The long-term vision for this architecture culminates in \pn{Universal MoSKA}. This vision is predicated on the maturation of \pn{Position-Independent KV Caching} \cite{EPIC2410}. Removing positional dependencies untethers KV chunks from their original context, turning them into modular, composable blocks of knowledge. A \pn{Universal MoSKA} infrastructure could then host a distributed network of nodes, each serving a distinct domain-specific cache cluster. A complex user query could then be serviced by dynamically composing a context from this universal library, pulling relevant knowledge chunks from multiple domains on demand. This represents an important step towards truly flexible and powerful knowledge-based AI systems.

\vspace{-4mm}
\section{Evaluation}
\renewcommand{\thesubsection}{\Alph{subsection}} 
We evaluated the proposed architecture through a detailed analytical model. This approach is a standard and effective methodology in computer architecture, and its validity for LLM inference is supported by recent high-fidelity simulators like \pn{LIFE}\cite{ForecastingLLM2508}. \pn{LIFE} demonstrates that performance models built upon fundamental hardware constraints—namely compute FLOPS and memory bandwidth—can accurately predict the throughput of complex LLM inference workloads on real hardware. 

Our experimental setup assumes a Llama 3.1 8B model using FP8 precision with 75\% sparsity for sparse attention. 
This assumption is consistent with prior studies \cite{MoBA2502, LongHeads2402}, which establish that lightweight routing can achieve high sparsity levels ($\geq$75\%) while preserving the model's task performance.
The hardware consists of two DGX H200 nodes. Each H200 GPU has 141GB of memory, 4.8TB/s of memory bandwidth, and 1979 TFLOPS of FP8 compute performance. We compare MoSKA against four state-of-the-art baselines: FlashAttention\cite{Dao2022FlashAttentionFA}, SGLang\cite{SGLang2312}, LongHeads\cite{LongHeads2402}, and ChunkAttention\cite{ChunkAttention2402}. 
As qualitatively positioned in Table~\ref{tab:tech_comparison}, MoSKA integrates a unique combination of architectural features—most notably Shared KV Attention and a disaggregated infrastructure—that distinguish it from the baselines used in our analysis.
The workload involves a large shared context (1M to 16M tokens) and a smaller unique context (64K tokens) per request. The workload assumes a target generation speed of 35 tokens/second for each request, reflecting a realistic service-level objective for long-sequence service.

\begin{figure}[t]
\centering
\includegraphics[width=\linewidth]{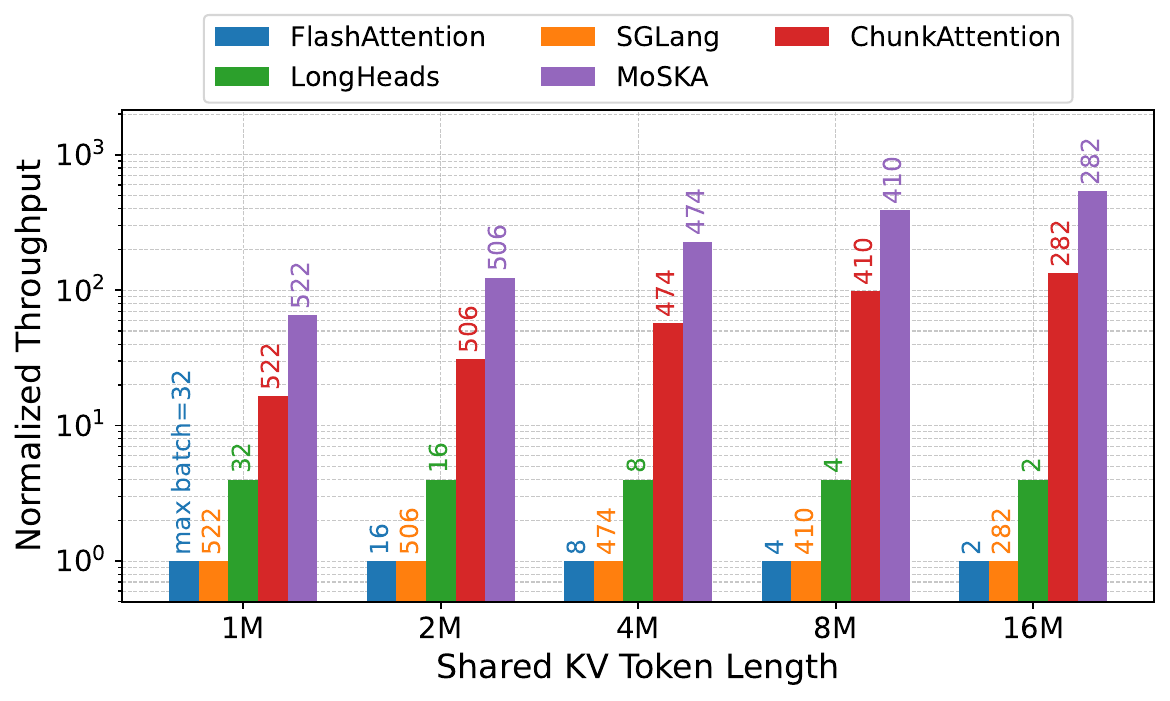}
\vspace{-6mm}
\caption{Batch scaling capability and normalized throughput.}
\label{fig:eval}
\vspace{-4mm}
\end{figure}

\vspace{-2mm}
\subsection{Batch Scaling Capability and Throughput}
This experiment evaluates how different architectures handle memory constraints and how that translates to system throughput. As shown in Fig.~\ref{fig:eval}, methods employing cache reuse and sharing—SGLang, ChunkAttention, and MoSKA—achieve substantially higher maximum batch sizes than FlashAttention and LongHeads by dramatically reducing the per-request memory footprint.

However, memory capacity is only one part of the story. The Fig.~\ref {fig:eval} reveals the performance advantage of \pn{Shared KV Attention}. Both ChunkAttention and \pn{MoSKA} significantly outperform other methods by transforming the attention on shared data into a compute-bound GEMM operation, thus solving the memory bandwidth bottleneck. \pn{MoSKA} consistently achieves the highest throughput, with a gain of up to 538.7$\times$ over the baseline. This superior performance is attributable to its ability to combine the high operational intensity of \pn{Shared KV Attention} with the computational savings of its \pn{MoE-inspired sparse attention}.

\vspace{-2mm}
\subsection{Resource Utilization in Disaggregated Infrastructure}
\begin{figure}[t]
\centering
\includegraphics[width=\columnwidth]{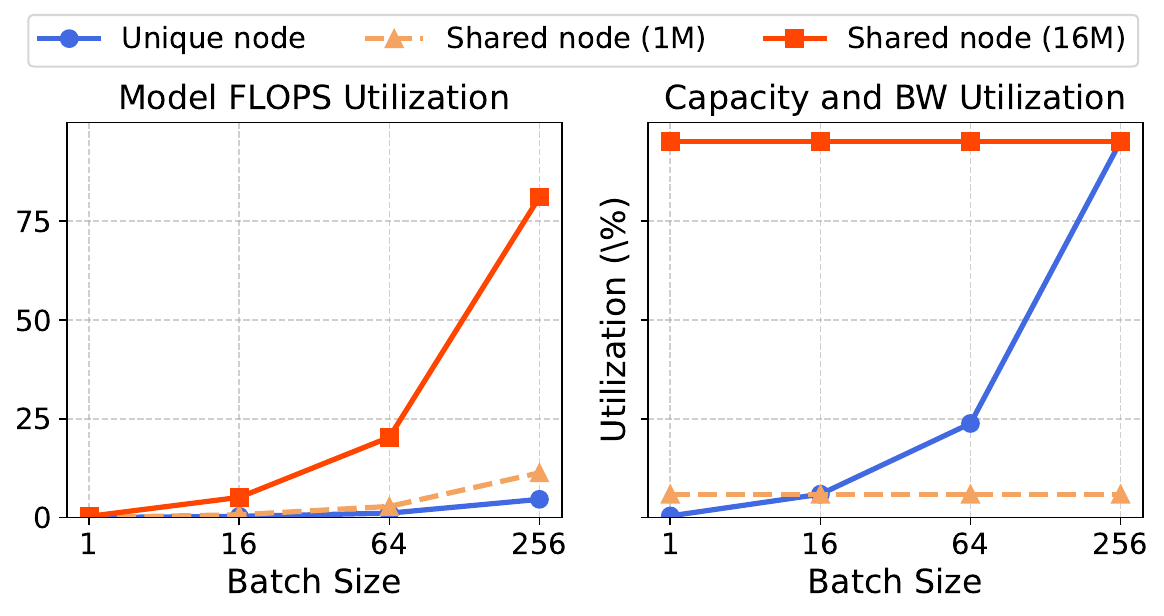}
\vspace{-6mm}
\caption{MFU and memory utilization of each node.}
\label{fig:gpu_util}
\vspace{-4mm}
\end{figure}

To validate our proposed disaggregated infrastructure, we analyzed the resource utilization of specialized nodes under increasing batch sizes, with a single DGX H200 node designated as a \pn{Unique KV} Node and another as a \pn{Shared KV} Node (Fig.~\ref{fig:gpu_util}).

The results clearly demonstrate the design's effectiveness. The Shared Node exhibits exceptional scalability; as the batch size scales to 256, its memory and bandwidth utilization remain minimal because the shared cache is loaded only once. Crucially, its Model FLOPS Utilization (MFU) scales almost linearly with the batch, reaching over 80\% for a 16M shared context. This confirms that Shared KV Attention successfully makes the workload compute-bound.

In stark contrast, the Unique Node’s capacity and bandwidth requirements scale linearly with the batch size, while MFU remains exceedingly low, a clear sign of its memory-bound nature. This result validates our design: by offloading the shared workload, the Unique Node is freed from this scalable but inefficient task and can be better optimized, for instance by co-locating \pn{FFN} computations to hide its inherent memory latency.

\vspace{-2mm}
\section{Conclusion and Future Work}
\pn{MoSKA}, a comprehensive architecture, is designed to systematically address the critical KV cache bottlenecks in long-sequence LLM inference. By transforming the attention on shared contexts from a memory-bound problem into a compute-bound one, \pn{MoSKA} achieves significant performance improvements, demonstrating up to a 538.7$\times$ throughput gain over baselines in our evaluation.

Future work will focus on the full-scale implementation of the \pn{MoSKA} architecture and its sophisticated scheduling algorithms, while advancing the \pn{Universal MoSKA} vision by contributing to research on position-independent transformers for dynamic knowledge composition.

\bibliographystyle{IEEEtran}
\bibliography{ref}

\vfill
\end{document}